\documentclass[runningheads,a4paper]{llncs}

\usepackage{amssymb}
\usepackage{graphicx}
\usepackage{comment}
\usepackage{array}
\usepackage{textcomp}
\usepackage{amsmath}
\usepackage[titlenumbered,ruled]{algorithm2e}
\usepackage{enumitem}

\usepackage[symbol]{footmisc}

\usepackage{xspace}

\def\etal{\textit{et al.}\xspace}
\def\ie{\textit{i.e.}~}

\def\vs{\emph{vs}.~}
 
\newcommand{\ra}[1]{\renewcommand{\arraystretch}{#1}}

\usepackage{color}

\newcommand{\JimSan}{Jim\'{e}nez-S\'{a}nchez}

\usepackage{booktabs}% http://ctan.org/pkg/booktabs
\usepackage{cite} % group citations
\usepackage[hidelinks]{hyperref} % http://ctan.org/pkg/hyperref

\usepackage{pifont} % http://ctan.org/pkg/pifont

\usepackage[labelfont=bf]{caption}
\usepackage{subcaption}

\usepackage{url}
\urldef{\mailsa}\path|{alfred.hofmann, ursula.barth, ingrid.haas, frank.holzwarth,|
\urldef{\mailsb}\path|anna.kramer, leonie.kunz, christine.reiss, nicole.sator,|
\urldef{\mailsc}\path|erika.siebert-cole, peter.strasser, lncs}@springer.com|    
\newcommand{\keywords}[1]{\par\addvspace\baselineskip
\noindent\keywordname\enspace\ignorespaces#1}

\begin{document}

\mainmatter  % start of an individual contribution

\title{Medical-based Deep Curriculum Learning \\ for Improved Fracture Classification}

\titlerunning{Medical-based Deep Curriculum Learning for Improved Classification}

\author{Amelia Jim\'{e}nez-S\'{a}nchez\inst{1}\href{https://orcid.org/0000-0001-7870-0603}{\includegraphics[scale=0.5]{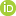}}, Diana Mateus\inst{2}\href{http://orcid.org/0000-0002-2252-8717}{\includegraphics[scale=0.5]{orcid.png}}, Sonja Kirchhoff\inst{3, 4}, \\ Chlodwig Kirchhoff\inst{4}, Peter Biberthaler\inst{4}, Nassir Navab\inst{5}, \\ Miguel A. Gonz\'{a}lez Ballester\inst{1,6}, Gemma Piella\inst{1}\href{https://orcid.org/0000-0001-5236-5819}{\includegraphics[scale=0.5]{orcid.png}}}

\authorrunning{A. \JimSan{} \etal{}}

\institute{
BCN MedTech, DTIC, Universitat Pompeu Fabra, Barcelona, Spain \\
\and 
Ecole Centrale de Nantes, LS2N, UMR CNRS 6004, Nantes, France \\
\and 
Institute of Clinical Radiology, LMU M{\"u}nchen, Munich, Germany \\
\and 
Department of Trauma Surgery, Klinikum rechts der Isar, \\ Technische Universit{\"a}t M{\"u}nchen, Munich, Germany \\
\and
Computer Aided Medical Procedures, Technische Universit{\"a}t M{\"u}nchen, Germany 
\and
ICREA, Barcelona, Spain
}

\toctitle{Lecture Notes in Computer Science}
\tocauthor{Authors' Instructions}
\maketitle

\begin{abstract}
Current deep-learning based methods do not easily integrate to clinical protocols, neither take full advantage of medical knowledge. In this work, we propose and compare several strategies relying on curriculum learning, to support the classification of proximal femur fracture from X-ray images, a challenging problem as reflected by existing intra- and inter-expert disagreement. Our strategies are derived from knowledge such as medical decision trees and inconsistencies in the annotations of multiple experts, which allows us to assign a degree of difficulty to each training sample. We demonstrate that if we start learning ``easy'' examples and move towards ``hard'', the model can reach a better performance, even with fewer data. The evaluation is performed on the classification of a clinical dataset of about 1000 X-ray images. Our results show that, compared to class-uniform and random strategies, the proposed medical knowledge-based curriculum, performs up to 15\% better in terms of accuracy, achieving the performance of experienced trauma surgeons.

\end{abstract}

\section{Introduction}
In a typical educational system, learning relies on a curriculum that introduces new concepts building upon previously acquired ones. The rationale behind, is that humans and animals learn better when information is presented in a meaningful way rather than randomly. Bringing these ideas from cognitive science, Elman~\cite{Elman1993} proposed the \textit{starting small} concept, to train neural networks that learn the grammatical structure of complex sentences. The networks were only able to solve the task when starting with a small sample size, highlighting the importance of systematic and gradual learning. Bengio~\etal~\cite{Bengio2009-curriculum} made the formal connection between machine learning and starting small, demonstrating a boost in performance by combining \textit{curriculum learning} (CL) with neural networks on two toy examples: shape recognition and language modeling. In the medical image analysis community, the idea of exploiting CL with deep learning has only recently been explored \cite{Havaei2016, Maicas2018-likeRadiologists, Tang2018-AttentionGuidedCL}. Even though these techniques have been successful in applications such as image segmentation or computer-aided diagnosis, they remain agnostic of clinical standards and medical protocols.  

Our goal is to fill the gap between machine learning algorithms and clinical practice by introducing medical knowledge integrated in a CL strategy. Current applications of CL to medical images focus on gradually increasing context in segmentation, whereas active learning approaches aim for reducing annotation efforts. Our focus is on the integration of knowledge, extracted from medical guidelines, directly from expert recommendations, or from ambiguities in their annotations, to ease the training of convolutional neural networks (CNNs). Similar in spirit, \cite{Tang2018-AttentionGuidedCL} integrates, as part of a more complex method for disease localization, a curriculum derived from clinical reports, extracted by natural language processing. 

\begin{figure}[t]
    \centering
    \includegraphics[width=1.\textwidth]{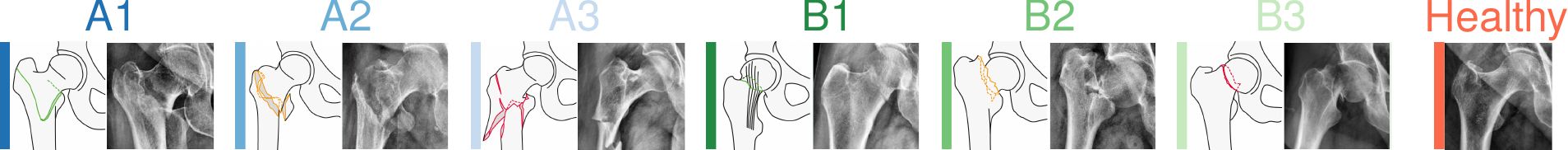}
    \caption{\textbf{AO standard and example radiographs.}}
    \label{fig:ao-examples}
\end{figure}

We restrict our study to the classification of proximal femur fractures according to the AO standard~\cite{Kellam2018}. Some example X-ray images with their corresponding category are depicted in Fig.~\ref{fig:ao-examples}. This kind of fracture represents a notable problem in our society, especially in the elderly population, having a direct socioeconomic repercussion \cite{moran2005early}. Early detection and classification of such fractures are essential for guiding appropriate treatment and intervention. However, several years of training are needed, and inter-reader agreement ranges between 66-71\% for trauma surgery residents and experienced trauma surgeons, respectively \cite{vanEmbden2010}. A similar classification problem was recently addressed in \cite{kazi2017AutomaticClassification}, where the focus was on the use of attention methods to improve fracture classification.

We explore the potential of curriculum learning to design three medical-based data schedulers that control the training sequence every epoch. Results on multi-label fracture classification show that by using our curriculum approaches, $F_1$-score can be improved up to 15\%, outperforming two baselines: without curriculum (random) and uniform-class reordering. Our proposed medical data schedulers, on restricted training data, outperform also the baselines having access to the entire training set. Furthermore, we reach a comparable performance to those of experienced trauma surgeons.

%%%%%%%%%%%%%%%%%%%%%%%%%%%%  Figure  %%%%%%%%%%%%%%%%%%%%%%%%%%%% 
%%%%%%%%%%%%%%%%%%%%%%%%%%%%%%%%%%%%%%%%%%%%%%%%%%%%%%%%%%%%%%%%%

\section{Related Work}
Curriculum learning's main hypothesis is that the order in which samples are presented to an iterative optimizer is important, as it can drastically change the outcome (\ie which local-minima is found). Bengio~\etal~\cite{Bengio2009-curriculum} showed that a significant improvement in the generalization of the machine learning algorithm could be achieved when samples were presented from ``easy'' to ``hard'', where the ``hardness'' of the samples was determined by some heuristic or a human expert. Similar to Bengio~\etal, we propose a series of heuristics to infer the hardness level of the training samples, but taking as guiding principles medical knowledge in the form of standards, such as the AO \cite{Kellam2018}, or disagreement between experts. 

There is little prior work in CL for medical image analysis. Maicas~\etal~\cite{Maicas2018-likeRadiologists} proposed to emulate how radiologists learn based on a set of increasing difficulty tasks. A combination of meta-learning and teacher-student curriculum learning allowed them to improve the selection of the tasks to achieve a boosted detection of malignancy cases in breast screening. For medical image segmentation, \cite{Havaei2016} also demonstrated that gradually increasing the difficulty of the task benefits the performance. A pseudo-curriculum approach was employed for segmentation of brain tumors and multiple sclerosis lesions in magnetic resonance (MR) images. Training starts from an easy scenario, where learning can be done from multi-modal MR images, and after a few warm-up epochs some of the modalities are randomly dropped. 

Novel variants of curriculum learning include self-paced learning, where the curriculum is automated. Such approach has been used to tackle imbalance in the segmentation of lung nodules in computed tomography images \cite{Jesson2018-CASED}. Instead of relying on prior medical knowledge, their approach updates the curriculum with respect to the model parameters' change, letting the learner focus on knowledge near the decision frontier, where examples are neither too easy nor too hard. Self-paced learning and active learning were used in \cite{Wang2018-DASL} to reduce annotation efforts. Though related, active learning has a different motivation than our work, as it focuses on retrieving examples from an unlabeled pool aiming to achieve a better performance with fewer labeled data. In our study, we restrict to the original concept of CL aiming to ease the learning process and improve the classification performance with the medical-based data schedulers. 

%%%%%%%%%%%%%%%%%%%%%%%%%%%%  Figure  %%%%%%%%%%%%%%%%%%%%%%%%%%%% 
\begin{figure}[t]
    \centering
    \includegraphics[width=1.\textwidth]{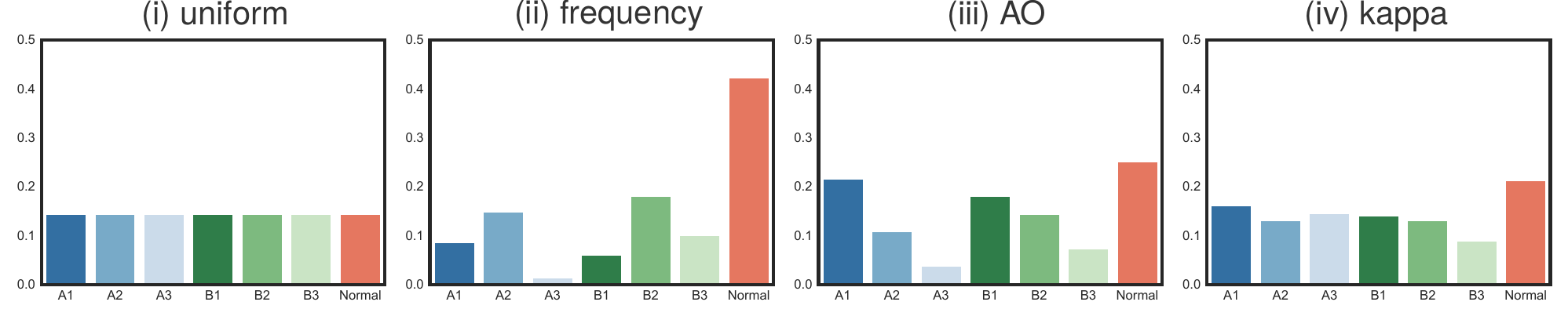}
    \captionsetup{singlelinecheck=off}
    \caption[empty]{\textbf{Initial probabilities $p^{(0)}$ for our medical-based curriculums:}
    \begin{enumerate}[label=\textbf{(\roman*})]
        \item \textbf{uniform}: samples are selected according to a uniform class-distribution.
        \item \textbf{frequency}: proportional to their original frequency in the dataset.
        \item \textbf{AO}: probabilities are assigned based on the difficulty of the AO classes.
        \item \textbf{kappa}: intra-rater Cohen's kappa coefficient to define the probabilities.
    \end{enumerate} }
    \label{fig:curriculums}
\end{figure}
%%%%%%%%%%%%%%%%%%%%%%%%%%%%%%%%%%%%%%%%%%%%%%%%%%%%%%%%%%%%%%%%%

\section{Methods}
We tackle multiclass image classification problems where an image $x_i$ needs to be assigned to a discrete class label $y_i \in \{y_1, y_2, \dots, y_M\}$. Let us consider the training set $\{\mathcal{X, Y}\}$ composed of $N$ element pairs, and assume training is performed in mini-batches of size $B$ for a total of $E$ epochs.
We address the problem by training a CNN with stochastic gradient descent (SGD) along with CL, favoring easier examples at the beginning of the training while solving the non-convex optimization in the long term. A curriculum $c \in \mathcal{C}$ induces a bias in the order of presenting samples to the optimizer. The bias reflects a notion of ``hardness'', which in this work depends upon different forms of prior medical knowledge. In practice, any of the curriculums is implemented by assigning a probability to each training pair, such that simpler cases have higher probabilities of being selected first. Over different rounds, the probabilities are updated according to a scheduler to reach a uniform distribution at the end. 

Initially, each image $x_i \in \mathcal{X}$ is assigned a curriculum probability $p_i^{(0)}$, here defined according to medical knowledge (see Eq.~\ref{eq:PCL} for practical definitions of $p_i^{(0)}$). At the beginning of every epoch, the training set $\{\mathcal{X, Y}\}$ is permuted to $\{\mathcal{X, Y}\}^{c}$ using a reordering function $f^{(e)}$. This mapping results from sampling the training set according to the probabilities at the current epoch $p^{(e)}$. Mini-batches are then formed from $\{\mathcal{X, Y}\}^{c}$ and the probabilities are decayed towards a uniform distribution \cite{Bengio2009-curriculum}, based on the following function \cite{Ren2018SelfPacedPC}: 
\begin{align}
q_i^{(e)} &= p_i^{(e-1)} \cdot \exp({-cn_i^2}/{10}) \quad \forall e > 0, \label{eq:decay1} \\
p_i^{(e)} &=  \dfrac{q_i^{(e)}}{{\sum_{i=1}^{N}q_i^{(e)}}},
\label{eq:decay2}
\end{align}
where $cn_i$ is a counter that is incremented when sample $i$ is selected.
The process for training a CNN with our medical curriculum data scheduler is summarized in Algorithm~\ref{alg:curriculum}.

%%%%%%%%%%%% Algorithm %%%%%%%%%%%%
\begin{algorithm}[bt]
\SetAlgoLined
\SetKwInOut{Input}{input}
\SetKwInOut{Output}{output}
\Input{$\mathcal{X}$ (X-ray images), $\mathcal{Y}$ (classification labels), $c \in \mathcal{C}$ (curriculum)  \\ 
$B$ (mini-batch size) , 
$E$ (expected training epochs)}
 \For{each epoch \textbf{e}}{ 
  \eIf{first epoch}{ 
    Define initial probabilities: $p_i^{(0)} = w_m^{c} $\; 
   }{
   Update probabilities with Eqs.~(\ref{eq:decay1}-\ref{eq:decay2})\; 
  }
  Get reordering function $f^{(e)}$ by sampling $ \{\mathcal{X},\mathcal{Y}\}$ according to $p^{(e)}$\\  
  Permute training set $f^{(e)}:\{\mathcal{X},\mathcal{Y}\} \mapsto \{\mathcal{X},\mathcal{Y}\}^{c} $ \;
   \For{each training round}{
    Get the \textbf{next} mini-batch from $\{\mathcal{X},\mathcal{Y}\}^{c}: \{x_b, y_b\}_{b=1}^{B} $\;
    Calculate cross-entropy loss $\mathcal{L}(y_b,\hat{y}_b)$\;
    Compute gradients and update model weights\;
    }
 }
 \caption{CNN with medical curriculum data scheduler}
 \label{alg:curriculum}
\end{algorithm}
%%%%%%%%%%%%%%%%%%%%%%%%%%%%%%%%%%%%

The proposed CL method is demonstrated on the classification of proximal femur fractures according to the AO system. In this standard, the first level of distinction differentiates fractures of type ``A'', located in the trochanteric region, from type ``B'' found in the femoral neck. Further subdivision of classes A and B depends on the specific location of the fracture and its morphology, \ie the number of fragments and their arrangement. We target the fine-grained classification to distinguish 6 types of fracture (A1-A3, B1-B3) plus the non-fracture case, \ie a 7-class problem, as shown in Figure~\ref{fig:ao-examples}.

The curriculum probabilities of each of the classes are given by:
\begin{equation} 
 p^{(0)}(y_i=m) = w_m^{c}, \\
\label{eq:PCL}
\end{equation}
where $m \in [1, 2, \dots, M]$ serves as index of the classes, and $w_m^{c}$ is defined according to $c \in \mathcal{C} = \{\text{uniform}, \text{frequency}, \text{AO}, \text{kappa}\}:$

\begin{itemize} 
    \item[$c$]: \textbf{uniform} (see Fig.~\ref{fig:curriculums}-(i)): all classes are treated equally, \ie,
    \begin{equation}
    w_m = 1/M. 
    \end{equation}
    \item[$c$]: \textbf{frequency} (see Fig.~\ref{fig:curriculums}-(ii)): classes are assigned a probability equal to their original frequency of appearance in the dataset, 
    \begin{equation}
        w_m = \dfrac{1}{N}\sum_{i=1}^{N} \delta_{y_i, m},
    \end{equation}
    where $\delta$ is the indicator function equal to one when $y_i=m$, and $0$ otherwise.
    \item[$c$]: \textbf{AO} (see Fig.~\ref{fig:curriculums}-(iii)): an experienced radiologist ranked the difficulty of the classes in the following order $v = [\text{A3}, \text{B3}, \text{A2}, \text{B2}, \text{B1}, \text{A1}, \text{non-fracture}]$ from hardest to easiest. As a naive approach, we consider the categories equally spaced and use the rank index $k$, such that:
    \begin{equation}
    w_m = \dfrac{k}{\sum_{m=1}^{M} m}.
    \end{equation}
    \item[$c$]: \textbf{kappa} (see Fig.~\ref{fig:curriculums}-(iv)): Cohen's kappa statistic is used to measure the agreement of clinical experts on the classification between two readings. Basically, kappa quantifies the ratio between the observed and chance agreement. Here, each class is assigned a probability proportional to the intra-reader agreement found by a committee of experts. 
\end{itemize}

\section{Experimental Validation}
\paragraph{\textbf{Dataset}.} Our in-house dataset consists of anonymized X-rays of the hip and pelvis collected at the trauma surgery department of the Rechts der Isar Hospital in Munich. The studies contain lateral view and anterior-posterior images. The latter, which involved both femora, were parted into two, resulting into additional non-fracture examples. The dataset consists of 327 type-A, 453 type-B fractures and 567 non-fracture cases. Subtypes of the fracture classes are highly unbalanced, reflecting the incidence of the different fracture types, as depicted in Fig.~\ref{fig:curriculums}-(ii). To address this problem, offline data augmentation techniques such as translation, scaling and rotation were used. The dataset was split patient-wise into three parts with the ratio 70\%\,:\,10\%\,:\,20\% to build respectively the training, validation and test sets. We employ a test distribution that is balanced between fracture type-A, type-B and non-fracture. All evaluations below are based on the weighted $F_1$-score, which takes into account the unbalanced class-distribution using as weights the support of each class.

Clinical experts provided along with the classification (based on the AO standard) a square region of interest (ROI) around the femur. ROIs were downsampled to $224 \times 224$ px to fit into the proposed architecture.

%%%%%%%%%%%%%%%%%%%%%%%%%%%%%%%%%%%%
%% Figure: Results
\begin{figure}[t]
    \centering
    \includegraphics[height=0.4\textwidth]{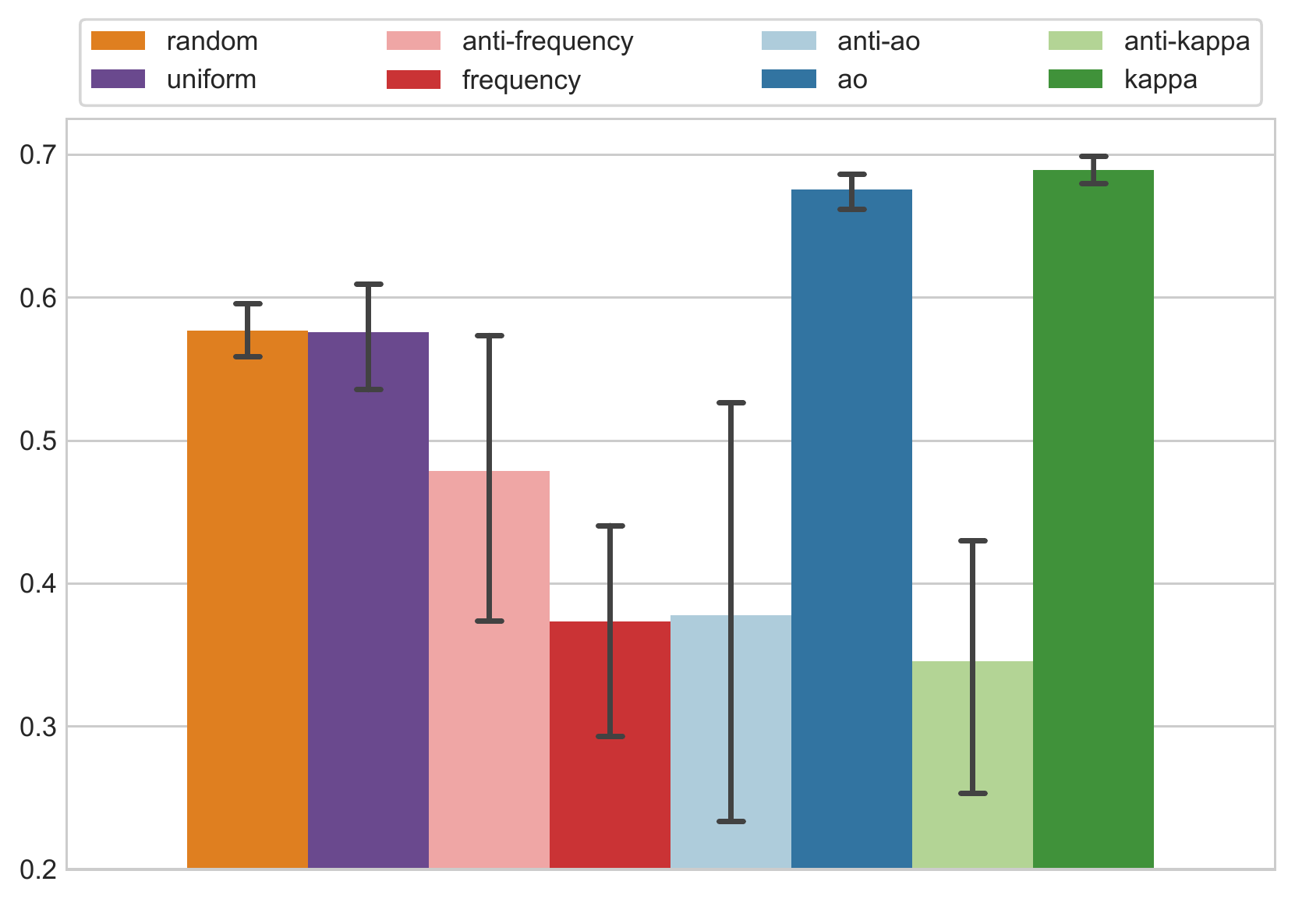}
    \caption{\textbf{Mean $F_1$-score and variance over 10 runs on the 7-way classification} of the different curriculum strategies, together with their corresponding anti-curriculum, and compared against random and uniform-curriculum.}
    \label{fig:barplot-results}
    \vspace{-1.5em}
\end{figure}
%%%%%%%%%%%%%%%%%%%%%%%%%%%%%%%%%%%%

\paragraph{\textbf{Implementation Details}.}
We used a ResNet-50~\cite{He2015-ResNet} pretrained on the Imagenet dataset. The architecture and curriculums were implemented with TensorFlow\footnote[2]{https://www.tensorflow.org/} and ran on an Nvidia Titan XP GPU. All models were run for 50 epochs with cross-entropy loss, in mini-batches of 64 samples and saved at minimum validation loss. SGD with a momentum of 0.9 was used as optimizer. Early stop was implemented if there was no improvement in the last 20 epochs. 70\% dropout was used in the fully-connected layer. Initial learning rate was set to $1 \times 10^{-3}$ and decayed by 10 every 15 epochs. 

\subsection{Results}
We evaluate our proposed medical curriculum data schedulers in which difficulty is gradually increased ($\mathcal{C}$: easy $\xrightarrow{}$ hard) by comparing them against two baseline approaches: random permutations and class-uniform reordering. Besides, we discuss their performance with respect to the opposite strategies in which difficulty is decreased, and we refer to as anti-curriculum (anti-$\mathcal{C}$: hard $\xrightarrow[]{}$ easy).

%%%%%%%%%%%%%%%%%%%%%%%%%%%%%%%%%%%%%%%%%%%%%%%%%%%%%%%%%%%%%%%%%%%%%%
%% Table: F1score results
\begin{table*}[t] 
\centering
\ra{1.2}
\caption{Classification results over 10 runs. The highlighted indices in bold correspond to the best two models.}
\begin{tabular}{@{}lccccccc@{}} \toprule
& \multicolumn{3}{c}{7 classes} & \phantom{abc} & \multicolumn{3}{c}{3 classes} \\
\cmidrule{2-4} \cmidrule{6-8} 
$F_1$-score & Mean & Median & SD & \phantom{abc} & Mean & Median & SD \\ \midrule
Random & 0.5662 & 0.5731 & 0.0423 & \phantom{abc} & 0.8063  & 0.8171 & 0.0337 \\
Uniform & 0.5757 & 0.5923 & 0.0590 & \phantom{abc} & 0.8011 & 0.7971 & 0.0399 \\
\midrule
AO & 0.6757 & {\bf 0.6783} & 0.0197 & \phantom{abc} & 0.8651 & {\bf 0.8657} & 0.0172 \\
Kappa & 0.6893 & {\bf 0.6900} & 0.0150 & \phantom{abc} & 0.8623 & {\bf 0.8657} & 0.0146\\
\midrule
AO - $60\%$ & 0.6325 & 0.6188 & 0.0302 & \phantom{abc} & 0.8457 & 0.8486 & 0.0191 \\
Kappa - $60\%$ \phantom{a} & 0.6352 & 0.6500 & 0.0398 & \phantom{abc} & 0.8446 & 0.8457 & 0.0222 \\
\bottomrule
\end{tabular}
\label{table:limited}
%\vspace{-1em}
\end{table*}
%%%%%%%%%%%%%%%%%%%%%%%%%%%%%%%%%%%%%%%%%%%%%%%%%%%%%%%%%%%%%%%%%%%%%%

Figure~\ref{fig:barplot-results} presents the results over 10 runs of the 7-class classification problem. Firstly, we find a similar performance between randomly shuffling the training data and learning with a uniform-curriculum, with a mean (median) $F_1$-score of 0.57 (0.57) and 0.58 (0.59), respectively. Secondly, we found that the sequence of the samples presented in each epoch has a significant effect in the classification, \ie there is a clear difference between curriculum and anti-curriculum strategies. Interestingly, our experiments suggest that, in the case of the frequency-curriculum, the easy scenario is the class-imbalance, which agrees with results reported in \cite{Wang2019DynamicCL}. The behaviour could also be related to the high imbalance in the original distribution and the offline augmentation. Finally, our two explicit medical AO- and kappa-curriculums boost the median $F_1$-score by approximately 15\% when compared against the baselines. The differences were found statistically significant (Suppl. Material). Moreover, our proposed schedulers help to reduce variance over the runs. 

By aggregating the posterior probability distribution obtained from our model, we can evaluate a 3-class problem, \ie transform the predictions to ``A'', ``B'' and ``non-fracture''. Although, we did not provide any supervision regarding the 3-class problem while training the CNN, we obtain a median $F_1$-score of 0.87 for AO- and kappa-curriculums, outperforming state-of-the-art results \cite{kazi2017AutomaticClassification}, and about 7\% better than random and uniform (0.82, 0.80). This means that mispredictions are usually within the same fracture type.

Our dataset size is typical for medical applications. An additional experiment was performed, under a restricted amount of balanced training data (only using 60\%), to investigate the performance of our medical-based data schedulers under reduced amounts of annotated data. Our AO- and kappa-curriculums performed even better than the baselines using all data (see Table~\ref{table:limited}).

Analyzing the training and validation loss curves while learning, we observed that random and uniform-curriculum converged smoothly and fast, whereas our proposed medical-based data schedulers were ``noisier'' in the first epochs. We hypothesize that the curriculum might lead to a better exploration of the weights during the first epochs.

\section{Conclusions} 
We have shown that the integration of medical knowledge is useful for the design of data schedulers by means of CL. Although we have focused on the AO standard and the multi-class proximal femur fracture problem, our work could be exploited in other applications where medical decision trees are available, such as grading malignancy of tumors, as well as whenever inter-expert agreement is available. As future work, we plan to explore the combination of our medical curriculum data schedulers with uncertainty of the model, and investigate which samples play a more significant role in the decision boundary.

\subsubsection*{Acknowledgments.} 
This project has received funding from the European Union’s Horizon 2020 research and innovation programme under the Marie Sk\l{}odowska-Curie grant agreement No. 713673 and by the Spanish Ministry of Economy [MDM-2015-0502]. A. \JimSan{} has received financial support through the ``la Caixa'' Foundation (ID Q5850017D), fellowship code: LCF/BQ/IN17/11620013. D. Mateus has received funding from Nantes M\'etropole and the European Regional Development, Pays de la Loire, under the Connect Talent scheme. Authors thank Nvidia for the donation of a GPU.

% splncs04: splncs03 citations in order of appearance instead of alphabetical
\bibliographystyle{splncs04}
\bibliography{biblio}

\begin{thebibliography}{10}
\providecommand{\url}[1]{\texttt{#1}}
\providecommand{\urlprefix}{URL }

\bibitem{Elman1993}
Elman, J.L.: Learning and development in neural networks: the importance of
  starting small. Cognition  48(1),  71--99 (jul 1993)

\bibitem{Bengio2009-curriculum}
Bengio, Y., Louradour, J., Collobert, R., Weston, J.: Curriculum learning. In:
  Proceedings of the 26th Annual International Conference on Machine Learning.
  pp. 41--48. ICML '09, ACM, New York, NY, USA (2009)

\bibitem{Havaei2016}
Havaei, M., Guizard, N., Chapados, N., Bengio, Y.: Hemis: Hetero-modal image
  segmentation. In: Ourselin, S., Joskowicz, L., Sabuncu, M.R., Unal, G.,
  Wells, W. (eds.) Medical Image Computing and Computer-Assisted Intervention
  -- MICCAI 2016. pp. 469--477. Springer International Publishing, Cham (2016)

\bibitem{Maicas2018-likeRadiologists}
Maicas, G., Bradley, A.P., Nascimento, J.C., Reid, I., Carneiro, G.: Training
  medical image analysis systems like radiologists. In: Frangi, A.F., Schnabel,
  J.A., Davatzikos, C., Alberola-L{\'o}pez, C., Fichtinger, G. (eds.) Medical
  Image Computing and Computer Assisted Intervention -- MICCAI 2018. pp.
  546--554. Springer International Publishing, Cham (2018)

\bibitem{Tang2018-AttentionGuidedCL}
Tang, Y., Wang, X., Harrison, A.P., Lu, L., Xiao, J., Summers, R.M.:
  Attention-guided curriculum learning for weakly supervised classification and
  localization of thoracic diseases on chest radiographs. In: MLMI@MICCAI
  (2018)

\bibitem{Kellam2018}
Kellam, J.F., Meinberg, E.G., Agel, J., Karam, M.D., Roberts, C.S.:
  Introduction. Journal of Orthopaedic Trauma  32,  S1--S10 (jan 2018)

\bibitem{moran2005early}
Moran, C.G., Wenn, R.T., Sikand, M., Taylor, A.M.: Early mortality after hip
  fracture: is delay before surgery important? JBJS  87(3),  483--489 (2005)

\bibitem{vanEmbden2010}
van Embden, D., Rhemrev, S., Meylaerts, S., Roukema, G.: The comparison of two
  classifications for trochanteric femur fractures: The {AO}/{ASIF}
  classification and the jensen classification. Injury  41(4),  377--381 (apr
  2010)

\bibitem{kazi2017AutomaticClassification}
Kazi, A., Albarqouni, S., Sanchez, A.J., Kirchhoff, S., Biberthaler, P., Navab,
  N., Mateus, D.: Automatic classification of proximal femur fractures based on
  attention models. In: Wang, Q., Shi, Y., Suk, H.I., Suzuki, K. (eds.) Machine
  Learning in Medical Imaging. pp. 70--78. Springer International Publishing,
  Cham (2017)

\bibitem{Jesson2018-CASED}
Jesson, A., Guizard, N., Ghalehjegh, S.H., Goblot, D., Soudan, F., Chapados,
  N.: Cased: Curriculum adaptive sampling for extreme data imbalance. In:
  Descoteaux, M., Maier-Hein, L., Franz, A., Jannin, P., Collins, D.L.,
  Duchesne, S. (eds.) Medical Image Computing and Computer Assisted
  Intervention − MICCAI 2017. pp. 639--646. Springer International
  Publishing, Cham (2017)

\bibitem{Wang2018-DASL}
Wang, W., Lu, Y., Wu, B., Chen, T., Chen, D.Z., Wu, J.: Deep active self-paced
  learning for accurate pulmonary nodule segmentation. In: Frangi, A.F.,
  Schnabel, J.A., Davatzikos, C., Alberola-L{\'o}pez, C., Fichtinger, G. (eds.)
  Medical Image Computing and Computer Assisted Intervention -- MICCAI 2018.
  pp. 723--731. Springer International Publishing, Cham (2018)

\bibitem{Ren2018SelfPacedPC}
Ren, Z., Dong, D., Li, H., Chen, C.: Self-paced prioritized curriculum learning
  with coverage penalty in deep reinforcement learning. IEEE Transactions on
  Neural Networks and Learning Systems  29,  2216--2226 (2018)

\bibitem{He2015-ResNet}
He, K., Zhang, X., Ren, S., Sun, J.: Deep residual learning for image
  recognition. 2016 IEEE Conference on Computer Vision and Pattern Recognition
  (CVPR) pp. 770--778 (2016)

\bibitem{Wang2019DynamicCL}
Wang, Y., Gan, W., Yang, J., Wu, W., Yan, J.: Dynamic curriculum learning for
  imbalanced data classification. In: Proceedings of the IEEE International
  Conference on Computer Vision. pp. 5017--5026 (2019)

\end{thebibliography}

\newpage
\section*{Supplementary Material}
Table~\ref{table:supp-f1score} presents the full results over 10 runs for random and curriculum-based strategies, for 7- and 3-class classification. These numbers correspond to the values represented in Fig.~\ref{fig:barplot-results}. 

The lower part of Table~\ref{table:supp-f1score} shows the results for the experiment with limited amount of training data (60\%) over 10 runs. In the paper only AO- and kappa-curriculums were presented because they are the most interesting cases, here we include as well random and uniform.
\vspace{-1.5em}

%%%%%%%%%%%%%%%%%%%%%%%%%%%%%%%%%%%%%%%%%%%%%%%%%%%%%%%%%%%%%%%%%%%%%%
%% Table: Full F1score results (10 runs)
\begin{table*}[h]  % bht
\centering
\ra{1.3}
\caption{ $F_1$-score. Mean, median and standard deviation over 10 runs. \\ The highlighted indices in bold correspond to the best two models.} 
\begin{tabular}{@{}lccccccc@{}} \toprule
& \multicolumn{3}{c}{7 classes} & \phantom{abc} & \multicolumn{3}{c}{3 classes} \\
\cmidrule{2-4} \cmidrule{6-8} 
F1score & Mean & Median & Std & \phantom{abc} & Mean & Median & Std \\ \midrule
Random & 0.5662 & 0.5731 & 0.0423 & \phantom{abc} & 0.8063  & 0.8171 & 0.0337 \\
Uniform & 0.5757 & 0.5923 & 0.0590 & \phantom{abc} & 0.8011 & 0.7971 & 0.0399 \\
\midrule
Anti-frequency & 0.4787 & 0.5521 & 0.1609 & \phantom{abc} & 0.6960 & 0.7829 & 0.1888 \\
Frequency & 0.3734 & 0.4190 & 0.1152 & \phantom{abc} & 0.5749 & 0.6000 & 0.0977 \\
Anti-AO & 0.3778 & 0.4149 & 0.2423 & \phantom{abc} & 0.6320 & 0.6743 & 0.2051 \\
AO & {\bf{0.6757}} & {\bf{0.6783}} & {\bf{0.0197}} & \phantom{abc} & {\bf{0.8651}} & {\bf{0.8657}} & {\bf{0.0172}} \\
Anti-kappa & 0.3456 & 0.4071 & 0.1514 & \phantom{abc} & 0.6046 & 0.6514 & 0.1328 \\
Kappa & {\bf{0.6893}} & {\bf{0.6900}} & {\bf{0.0150}} & \phantom{abc} & {\bf{0.8623}} & {\bf{0.8657}} & {\bf{0.0146}}\\
\midrule
\midrule
Random-60\% & 0.5363 & 0.5330 & 0.0393 & \phantom{abc} & 0.7669 & 0.7600 & 0.0237 \\ 
Uniform-60\% &  0.5420 & 0.5625 & 0.0373 & \phantom{abc} & 0.7949 & 0.7971 & 0.0248 \\ 
AO-60\% &  0.6325 & 0.6188 & 0.0302 & \phantom{abc} & 0.8457 & 0.8486 & 0.0191 \\ 
Kappa-60\% &  0.6352 & 0.6500 & 0.0398 & \phantom{abc} & 0.8446 & 0.8457 & 0.0222\\
\bottomrule
\end{tabular}
\label{table:supp-f1score}
\end{table*}
%%%%%%%%%%%%%%%%%%%%%%%%%%%%%%%%%%%%%%%%%%%%%%%%%%%%%%%%%%%%%%%%%%%%%%

\vspace{-1.5em}
Table~\ref{table:pvalues} presents the p-values obtained from running a significance pair t-test comparing AO- and kappa-curriculums \vs random, uniform and their respective anti-curriculum strategy.

\begin{comment}
Figure~\ref{fig:barplot-3class} depicts the classification performance on the 3-class problem.
Figure~\ref{fig:confusion-matrices} visualizes the confusion matrices of the best two models for the fine-grained classification.
Figure~\ref{fig:train-val-curves} shows the train \vs validation cross-entropy loss during the first 20 epochs of training.
\end{comment}

%%%%%%%%%%%%%%%%%%%%%%%%%%%%%%%%%%%%%%%%%%%%%%%%%%%%%%%%%%%%%%%%%%%%%%
%% Table: pvalues results (10 runs)
\begin{table*}[h]  % bht
\centering
\caption{p-values to test statistical significance of the 10 runs with the whole training set (100\%).}
\ra{1.5}
\begin{tabular}{@{}lccc@{}} \toprule
p-value & Random & Uniform & Anti-curriculum  \\ \midrule
AO & $1.96 \times 10^{-7}$ & $1.38 \times 10^{-4}$ & $1.73 \times 10^{-3}$ \\
Kappa & $1.08 \times 10^{-8}$ & $2.61 \times 10^{-5}$ & $2.39 \times 10^{-6}$  \\
\bottomrule
\end{tabular}
\label{table:pvalues}
\end{table*}
%%%%%%%%%%%%%%%%%%%%%%%%%%%%%%%%%%%%%%%%%%%%%%%%%%%%%%%%%%%%%%%%%%%%%%

\end{document}